\documentclass[letterpaper, 10 pt, conference]{ieeeconf}  
\AtBeginDocument{%
  }
\usepackage{graphicx}
\usepackage{amsfonts}
\usepackage{amssymb}
\usepackage[linesnumbered,ruled,vlined]{algorithm2e}
\usepackage{amsmath}
\usepackage{multirow}  
\usepackage{adjustbox}
\usepackage{booktabs}
\usepackage{siunitx}
\IEEEoverridecommandlockouts                              
\title{\LARGE \bf
Towards Robust LiDAR Localization: Deep Learning-based Uncertainty Estimation
}

\author{Minoo Dolatabadi$^{1,\dagger}$, Fardin Ayar$^{1,\dagger}$, Ehsan Javanmardi$^{2}$, Manabu Tsukada$^{2}$, Mahdi Javanmardi$^{1,*}$
\thanks{$^\dagger$ Minoo Dolatabadi and Fardin Ayar contributed equally to this work.}%
\thanks{$^{1}$Department of Computer Engineering, Amirkabir University of Technology, Tehran, Iran.}%
\thanks{$^{2}$Graduate School of Information Science and Technology, The University of Tokyo, Tokyo, Japan.}%
\thanks{$^{*}$Corresponding Author: Mahdi Javanmardi, \texttt{mjavan@aut.ac.ir}.}}%
\begin{document}

\maketitle
\thispagestyle{empty}
\pagestyle{empty}

\begin{abstract}
LiDAR-based localization and SLAM often rely on iterative matching algorithms, particularly the Iterative Closest Point (ICP) algorithm, to align sensor data with pre-existing maps or previous scans. However, ICP is prone to errors in featureless environments and dynamic scenes, leading to inaccurate pose estimation. Accurately predicting the uncertainty associated with ICP is crucial for robust state estimation but remains challenging, as existing approaches often rely on handcrafted models or simplified assumptions. Moreover, a few deep learning-based methods for localizability estimation either depend on a pre-built map—which may not always be available—or provide a binary classification of localizable versus non-localizable, which fails to properly model uncertainty.
In this work, we propose a data-driven framework that leverages deep learning to estimate the registration error covariance of ICP before matching, even in the absence of a reference map. By associating each LiDAR scan with a reliable 6-DoF error covariance estimate, our method enables seamless integration of ICP within Kalman filtering, enhancing localization accuracy and robustness. Extensive experiments on the KITTI dataset demonstrate the effectiveness of our approach, showing that it accurately predicts covariance and, when applied to localization using a pre-built map or SLAM, reduces localization errors and improves robustness.
\end{abstract}

\section{INTRODUCTION}
In recent years, autonomous vehicles have become an integral component of intelligent transportation systems, driving continuous research to push the boundaries of their capabilities. A key requirement for autonomous driving is achieving precise self-localization at the centimeter level \cite{javanmardi2019autonomous}. The Global Navigation Satellite System (GNSS) is a cost-effective and widely used method for vehicle localization. Although GNSS delivers reliable positioning in open-sky environments, its accuracy is significantly compromised in urban settings due to factors such as signal blockage, non-line-of-sight (NLOS) conditions, and multipath effects \cite{ellul2016impact}. To address these challenges, vision-based approaches—particularly those leveraging LiDAR—have been proposed as alternative or complementary solutions \cite{seo2014tracking}. In these systems, map matching techniques like the Iterative Closest Point (ICP) algorithm are frequently employed to align sensor data with pre-existing maps, thereby enhancing localization precision \cite{charroud2024localization}.

However, vision-based localization methods—like other relative localization techniques—are susceptible to error accumulation, where minor errors can progressively lead to significant drift over time. This phenomenon is observed in both Simultaneous Localization and Mapping (SLAM) and pre-built map-based localization approaches, although it tends to be more pronounced in SLAM \cite{hao2023global}. In the context of ICP, previous work has demonstrated that factors such as featureless environments (e.g., tunnels) and the presence of dynamic objects can adversely affect the matching accuracy \cite{tuna2023x}.

State estimation methods, such as Kalman filtering, are commonly employed to mitigate these problems. Yet, these techniques depend on an accurate error model (often represented in the simplest form by an error covariance matrix), which is challenging to determine for matching algorithms 
\cite{landry2019cello}.

\begin{figure}[t]
\centering
  \includegraphics[width=\linewidth]{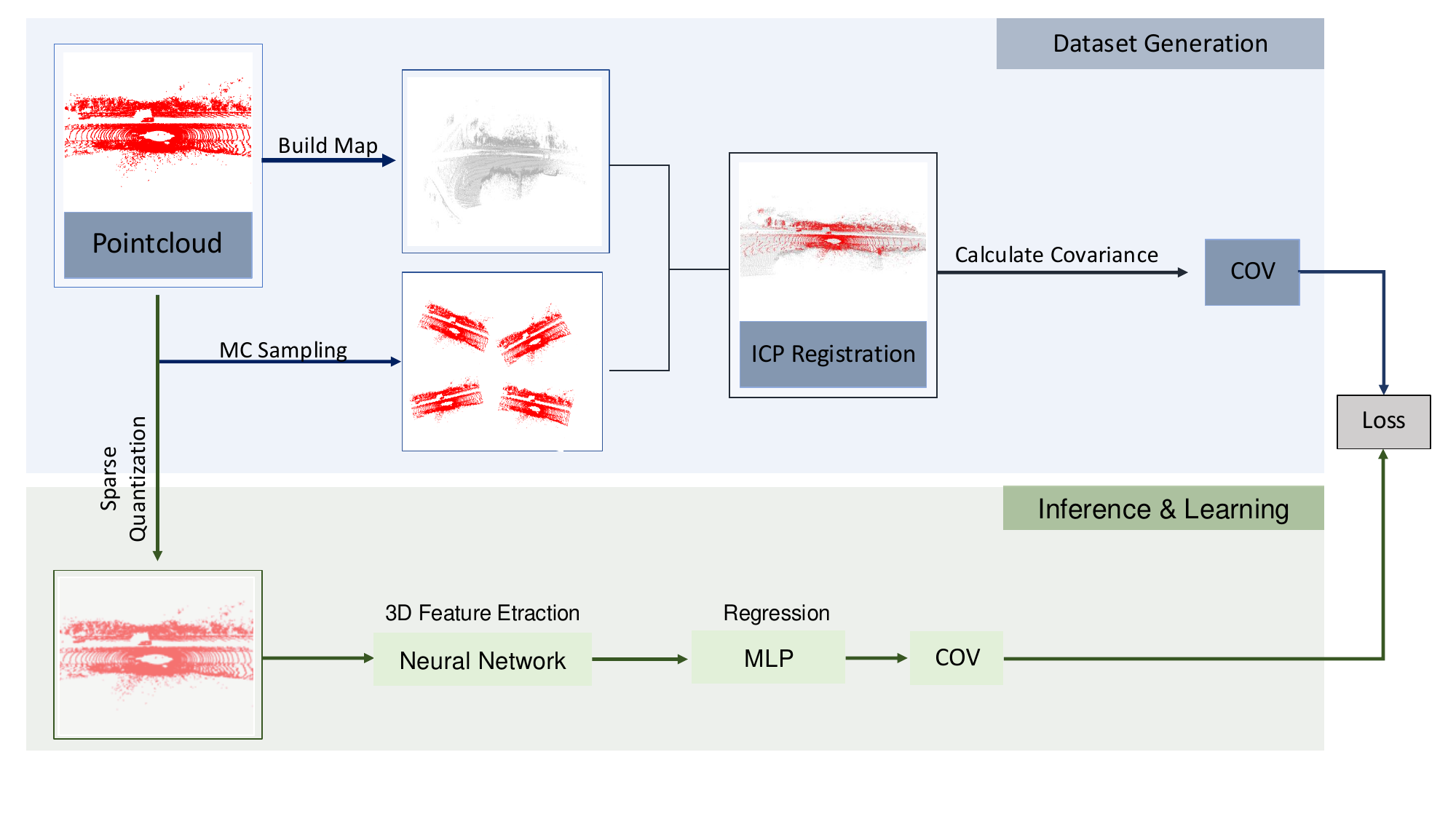}
  \caption[Proposed Approach Overview]{Overview of the proposed approach. In the upper section, a map is built from the input point cloud, followed by ICP matching to generate covariance for each point cloud. In the lower section, the input point cloud from the LiDAR sensor is processed through a neural network. The entire network, including feature extraction and regression, is trained end-to-end.}
  \label{our_net}
\end{figure}

To address this limitation, several studies have proposed data-driven approaches to predict ICP error, either using classification-based methods \cite{nubert2022learning} or by directly estimating the covariance \cite{landry2019cello}.

Specifically, we introduce a data-driven framework that predicts the full six-degree-of-freedom (6-DoF) ICP registration covariance from a \emph{single} LiDAR scan \emph{prior} to correspondence search and ICP refinement. The output is a symmetric positive-definite (SPD) $6\times6$ covariance on $SE(3)$ suitable for probabilistic fusion. An overview of the pipeline is shown in Fig.~\ref{our_net}.

\noindent\textbf{Contributions.}
\begin{itemize}
    \item \textbf{Pre-ICP, per-scan $SE(3)$ covariance.} We predict the full $6\times6$ registration covariance directly from a single LiDAR scan, capturing translation–rotation coupling before ICP is run. Per-scan training targets are obtained by estimating empirical covariances from Monte Carlo ICP under randomized initializations.
    \item \textbf{Kalman fusion and evaluation.} Each scan is paired with a reliable 6-DoF covariance, used as the measurement noise in a standard Kalman filter. This improves pose accuracy over fixed or heuristic covariances on the evaluated sequences.
    \item \textbf{Practicality.} The model does not require a pre-built map at inference, supporting both SLAM and map-based operation.
\end{itemize}



\section{RELATED WORKS}
As noted earlier, LiDAR-based localization typically registers scans with ICP \cite{besl1992method}\cite{low2004linear}; however, ICP can be unreliable in feature-poor or geometrically degenerate scenes \cite{tuna2023x}.  Predicting how well a LiDAR scan can be aligned before or during ICP is a crucial first step toward robust localization. Existing research in this area falls broadly into two complementary groups. i) \emph{Analytic Methods}, which use geometric and optimization diagnostics (e.g., scene structure, Jacobian conditioning, eigenvalue spectra) to detect degeneracy, and (ii) \emph{Learning-Based Methods}, which replace fixed heuristics with data-driven models that infer reliability and uncertainty directly from raw sensor data—yielding scalar or binary risk/localizability scores and learned covariance predictions (full $6\times 6$ pose uncertainty). This trajectory reflects a shift from explicit, model-based indicators to learned predictors of both risk and covariance; we review both in turn.
\subsection{Analytic Methods}
Classical approaches detect ICP failure modes by analyzing the registration cost and its local curvature. In practice, two strategies are common: (i) geometric or observability checks—e.g., looking for large planar areas or symmetric structures in the point cloud to flag ICP degeneracy \cite{gelfand2003geometrically}\cite{kwok2016improvements}—and (ii) curvature-based tests that inspect the Hessian’s eigenvalues or the ICP Jacobian’s condition number to reveal poorly constrained directions\cite{rong2016detection}\cite{ebadi2021dare}. While these techniques improve alignment stability, they typically require both the source and target scans and rely on heuristic thresholds, thereby limiting robustness in dynamic or sparse scenes. More recent variants address degeneracy like X-ICP \cite{tuna2023x} normalizes constraints to expose rank loss, and DARE-SLAM \cite{ebadi2021dare} handles perceptual aliasing; however, these improvements are applied after the fact rather than predicting issues upfront.
\subsection{Learning-Based Methods}
To overcome the above limitations, researchers have explored data-driven predictors of localization before or alongside alignment, as well as learned models of registration uncertainty. 
Prior work divides into two complementary strands. The first predicts or detects alignment risk—often before or during ICP—by measuring scan overlap or scene “localizability”, yielding a scalar or binary failure indicator. The second estimates the full $6\times6$ pose covariance of the ICP result, providing probabilistic uncertainty for Kalman filtering and other sensor-fusion methods. In practice, risk detection helps prevent catastrophic misalignments, while covariance estimation supplies the detailed uncertainty needed for consistent downstream state estimation. We review these strands in turn.
\subsubsection{Risk/Localizability Prediction}
Some learning-based methods predict pre-alignment localizability (risk) directly from raw LiDAR scans, enabling early detection of ICP failures. Nobili et al. \cite{nobili2018predicting} train an SVM classifier on
scan overlap and LiDAR intensity features to predict alignment failure probabilities. Similarly, Adolfsson
et al. \cite{adolfsson2021coral} quantify entropy change between scans to detect poor alignments. These methods produce binary
or scalar indicators of ICP reliability rather than a full uncertainty measure. More recent deep learning
approaches go further: Chen et al. \cite{chen2022overlapnet} propose OverlapNet, a Siamese network estimating overlap between
LiDAR scans , effectively predicting \emph{matchability}. SuperLoc \cite{zhao2024superloc} integrates LiDAR and
inertial data with a neural alignment risk predictor, enabling early detection of localization failures in
feature-scarce environments. Nubert et al. \cite{nubert2022learning} present a learning-based localizability estimator that
outputs a degeneracy score for LiDAR localization. In parallel, multisensor frameworks have been developed to mitigate degeneracy on the fly.
For instance, Xie et al. \cite{xu2024selective} introduce
a Selective Kalman Filter that fuses camera and LiDAR data only when the LiDAR-only SLAM is predicted to
be degenerate. By adaptively injecting visual constraints when needed, their system improves
robustness during ICP failure cases. 

In addition to ICP-based approaches, NDT-based localization has been systematically studied. Javanmardi et al. \cite{javanmardi2018factors} examined factors such as feature completeness and environmental structure to predict localization errors, demonstrating that map accuracy alone is not enough for precise localization. Their analysis successfully modeled urban scene errors based on these map features. Endo et al. \cite{endo2021analysis} further explored how dynamic elements, such as moving vehicles, influence localization accuracy, providing insights into improving NDT-based localization in real-world environments.
These advancements in both ICP and NDT-based systems highlight the transformative potential of data-driven methods in enhancing alignment and localization accuracy.
\subsubsection{Covariance Prediction}
Learning covariance with deep networks is challenging: the estimator must output a valid symmetric positive semidefinite matrix that captures both marginal scales and cross-correlations, while remaining numerically stable for real-time use. In related estimation tasks, Liu et al. \cite{liu2018deep} introduced a neural network
framework that predicts sensor measurement covariance,
offering real-time adaptability. Additionally, Fang et al. \cite{fang2021cnn}
applied deep learning to predict covariance matrices for
position estimation, ensuring semi-positive definiteness using
Cholesky decomposition.
In ICP-based localization, early work by Censi derived a closed-form covariance expression for ICP, laying the foundation for uncertainty analysis but relying on small-noise assumptions and linearization around the converged solution \cite{censi2007accurate}. Building on this idea, CELLO and CELLO-3D propagate geometric error sources to compute registration covariance in real time \cite{vega2013cello,landry2019cello}; however, like other analytic or feature-engineered methods, they still produce covariance only \emph{after} alignment and can be sensitive to poor initialization or scene structure. More recently, De Maio and Lacroix learn a Bayesian model of the ICP \emph{alignment error after registration}, yielding a data-driven, post-hoc covariance for the final solution \cite{de2022deep}.

%


\noindent\textbf{Discussion.}
We address the missing \emph{pre-ICP} regime by estimating a full $6\times6$ SE(3) covariance \emph{before} any correspondence search, from a single LiDAR scan—map-free and threshold-free. In contrast to methods that output only binary/scalar risk cues or rely on external degeneracy detectors \cite{nubert2022learning,xu2024selective}, our model infers degeneracy directly from the scan and returns a continuous, symmetric positive-definite covariance that captures both scale and directionality, making it directly usable in probabilistic filters. Post-registration estimators (e.g., \cite{de2022deep}) recover uncertainty \emph{after} alignment; our one-shot, data-driven predictor provides it \emph{a priori}. We train on Monte-Carlo–aligned covariance targets using modern point-cloud backbones (PointNet++, Cylinder3D), and analyses of ICP convergence behavior.

\section{Proposed Approach}

Given a 3D LiDAR point cloud scan \( s_k \in \mathbb{R}^{n_k \times 3} \), our objective is to predict a \(6 \times 6\) covariance matrix \( \Sigma_k \) that quantifies the uncertainty in the estimated relative pose if this point was matched against a reference map. In this work, we address two scenarios: one in which a pre-built map is available, and another where a local map generated via SLAM is used. 

The diagonal elements of the covariance matrix are denoted by
\[
\mathbf{c}_k = \operatorname{diag}\!\bigl(\boldsymbol{\Sigma}_k)=\bigl(\sigma^2_{t_x},\sigma^2_{t_y},\sigma^2_{t_z},\sigma^2_{\omega_x},\sigma^2_{\omega_y},\sigma^2_{\omega_z}\bigr)^\top,
\]
where the first three entries are translational variances (m\(^2\)) and the last three are rotational variances (rad\(^2\) or degree\(^2\)) about the \(x/y/z\) axes.

our method is built upon two key components:
\begin{enumerate}
    \item \textbf{Monte Carlo-based Dataset Generation:} This component creates a dataset from pre-built maps by generating ground truth \(6 \times 6\) error covariance matrices, which capture the uncertainty across various localization dimensions.
    \item \textbf{Deep Neural Network Framework:} This component is designed and trained to predict the error covariance matrix from the LiDAR scans. The predicted covariance, with its interpretable diagonal \(\mathbf{c}_k\), directly informs on the uncertainty of the localization estimate.
\end{enumerate}

An overview of the proposed method is presented in Figure \ref{our_net}.

\subsection{ICP and Its Uncertainty}

We utilize the Lie group \(SE(3)\) for rigid body transformations, which combines rotation and translation in three-dimensional space. The transformation matrix \(\mathbf{T} \in SE(3)\) can be expressed as:
\begin{equation}
    \mathbf{T} = \begin{pmatrix}
        \mathbf{R} & \mathbf{t} \\
        \mathbf{0}^\top & 1
    \end{pmatrix},
\end{equation}
where \(\mathbf{R} \in SO(3)\) is a rotation matrix and \(\mathbf{t} \in \mathbb{R}^3\) is a translation vector. To handle transformations analytically, it's beneficial to map \(\mathbf{T}\) to its corresponding Lie algebra \(\mathfrak{se}(3)\) using the logarithm map:
\begin{equation}
    \boldsymbol{\tau} = \log(\mathbf{T}) \in \mathfrak{se}(3).
\end{equation}

The vector \(\boldsymbol{\tau} \in \mathbb{R}^6\) comprises the translational and rotational components:
\begin{equation}
  \boldsymbol{\tau} = \begin{pmatrix}
    \mathbf{u} \\
    \boldsymbol{\omega}
  \end{pmatrix},
\end{equation}
where \(\mathbf{u} \in \mathbb{R}^3\) represents translation, and \(\boldsymbol{\omega} \in \mathbb{R}^3\) denotes the rotation in angle-axis form. This representation facilitates the expression of uncertainties in transformations as a covariance matrix \(\mathbf{\Sigma} \in \mathbb{R}^{6 \times 6}\):
\begin{equation}
    \mathbf{\Sigma} = \begin{pmatrix}
        \mathbf{\Sigma}_{\mathbf{u}\mathbf{u}} & \mathbf{\Sigma}_{\mathbf{u}\boldsymbol{\omega}} \\
        \mathbf{\Sigma}_{\boldsymbol{\omega}\mathbf{u}} & \mathbf{\Sigma}_{\boldsymbol{\omega}\boldsymbol{\omega}}
    \end{pmatrix},
\end{equation}
where \(\mathbf{\Sigma}_{\mathbf{u}\mathbf{u}}\) and \(\mathbf{\Sigma}_{\boldsymbol{\omega}\boldsymbol{\omega}}\) are the covariance matrices for translation and rotation, respectively, and \(\mathbf{\Sigma}_{\mathbf{u}\boldsymbol{\omega}}\) captures the correlation between them.

In practical applications, an initial transformation \( \mathbf{T} \) serves as a prior, often derived from odometry or other sensors. The Iterative Closest Point (ICP) algorithm refines this to obtain an estimated transformation \( \widehat{\boldsymbol{T}} \):
\begin{equation}
   \widehat{\boldsymbol{T}} = \text{ICP}(\mathbf{P}, \mathbf{Q}, \mathbf{T}), 
\end{equation}
where \( \mathbf{P} \) and \( \mathbf{Q} \) are the source and target point clouds, respectively. The uncertainty associated with \( \widehat{\boldsymbol{T}} \) is modeled as:
\begin{equation}
    \widehat{\boldsymbol{T}} = \exp(\boldsymbol{\xi}) \overline{\boldsymbol{T}}, \quad \boldsymbol{\xi} \sim \mathcal{N}(\mathbf{0}, \mathbf{\Sigma}),
\end{equation}
assuming the perturbation \( \boldsymbol{\xi} \) follows a Gaussian distribution with zero mean and covariance \( \mathbf{\Sigma} \) and \(\overline{\boldsymbol{T}}\) being the ground truth transformations.

To summarize, our goal is to learn a function 
\(
f_{\theta}: s_k \in \mathbb{R}^{n_k \times 3} \rightarrow \mathbf{\Sigma}_k \in \mathbb{R}^{6 \times 6},
\)
which is a deep neural network parameterized by \(\theta\) that predicts an error covariance matrix of the ICP algorithm, given only a scanned LiDAR point cloud.

\subsection{Dataset Generation}
To train the deep neural network for predicting the error covariance matrix in localization, we use a dataset generation process based on Monte Carlo sampling like \cite{nubert2022learning}. This process addresses the sensitivity of the ICP algorithm to its initial guess by repeatedly registering each point cloud under varying perturbations. Following the previous works \cite{nubert2022learning}, the overall procedure for each point cloud \( s_k \) is as follows:

\begin{enumerate}
    \item \textbf{Map Construction:} \\
    A map is constructed by aggregating a sequence of point clouds—specifically, \(x\) preceding and \(y\) subsequent scans. Each scan is transformed from its local coordinate frame into the map’s coordinate system using ground truth transformations. Notably, when estimating error covariance for SLAM, only preceding scans are used (i.e., \(y=0\)). This sequential accumulation progressively constructs an accurate representation of the environment. In cases where ground truth data are unavailable, mapping methods such as LOAM \cite{zhang2014loam} can be employed as an alternative.

    \item \textbf{Data Reduction:} \\
    To mitigate computational load and memory requirements, both the constructed map and the reference point cloud are downsampled via voxel filtering. In our implementation, the map is downsampled with a voxel size of 1 meter, while the reference point cloud is downsampled at a finer resolution of 0.1 meter. 

    \item \textbf{ICP Matching:} \\
    For each \( i \in \{0,\ldots,N\} \), we create an initial transformation \( \mathbf{T} \) (see Eq. 5) assuming it follows the model:
    \begin{equation}
        \mathbf{T}_i = \exp(\boldsymbol{\xi}_o) \, \overline{\mathbf{T}}_k,
    \end{equation}
    where \(\boldsymbol{\xi}_o \sim \mathcal{N}(\mathbf{0}, \mathcal{O})\) represents a perturbation around the ground truth transformation \(\overline{\mathbf{T}}_k\). The \(6 \times 6\) covariance matrix \(\mathcal{O}\) is a predefined hyperparameter representing the real-world uncertainty of the initial transformation (see Sec. IV). The exponential map converts \(\boldsymbol{\xi}_o\) into a valid transformation in \( SE(3) \).
    
    After applying the point-to-plane ICP, the refined transformation \(\widehat{\mathbf{T}}_i\) is obtained for the \(i\)-th sampled initial transformation.

    \item \textbf{Error Calculation and Covariance Estimation:} \\
    For each \(\widehat{\mathbf{T}}_i\), the registration error is quantified by:
    \begin{equation}
         \boldsymbol{\xi}_i = \log \left(\overline{\boldsymbol{T}}_k^{-1} \, \widehat{\boldsymbol{T}}_i\right),
    \end{equation}
    where the logarithm map transforms the error into its vector representation in \(\mathfrak{se}(3)\). After repeating the registration process for \( N \) samples, the covariance matrix is estimated by:
    \begin{equation}
        \boldsymbol{Y}_k = \frac{1}{n-1} \sum_{i=1}^{n} \boldsymbol{\xi}_i \boldsymbol{\xi}_i^{\top},
    \end{equation}
    where \(\boldsymbol{Y}_k \in \mathbb{R}^{6 \times 6}\) represents the error covariance matrix for the \( k \)-th point cloud.
\end{enumerate}

\bigskip
 The complete pseudocode for the dataset generation process is provided in Algorithm~\ref{alg-dataset}.

\begin{algorithm}[!ht]
    \caption{Dataset-Generation}\label{alg-dataset}
    \DontPrintSemicolon
    \LinesNumbered
    \KwIn{Set of point clouds}
    \KwOut{Covariance matrices $\boldsymbol{Y}_k$ for all $k$}
    
    \For{each point cloud $s_k$}{
        \For{$i \gets 1$ \KwTo $n$}{
            Construct map using $x$ previous and $y$ subsequent point clouds\;
            Transform point clouds from local to global coordinates using ground truth \;
            Initialize $\boldsymbol{\xi_o} \sim \mathcal{N}(\mathbf{0}, \mathcal{O})$\;
            Compute $\boldsymbol{T}_i = \exp(\boldsymbol{\xi}) \overline{\boldsymbol{T}}_k$\;
            Apply ICP to align input point cloud with map, obtain estimate $\widehat{\boldsymbol{T}}_i$\;
            Compute $\boldsymbol{\xi}_i = \log\left(\overline{\boldsymbol{T}}_k^{-1} \widehat{\boldsymbol{T}}_i\right)$\;
        }
        Compute covariance matrix $\boldsymbol{Y}_k = \dfrac{1}{n-1} \sum_i \boldsymbol{\xi}_i \boldsymbol{\xi}_i^{\top}$\;
    }
    \Return Covariance matrices $\boldsymbol{Y}_k$ for all $k$\;
\end{algorithm}

\subsection{Deep Learning for Predicting Error Covariance Matrix}
Our goal is to optimize a deep regression network \( f_{\theta} \) that predicts error covariance matrices from features extracted from the LiDAR scan \( s_k \). In this section, we first describe the network architecture, followed by our data augmentation strategy, and finally the loss function formulation. 
\subsubsection{Network Architecture}
We employ state-of-the-art architectures, Cylinder3D \cite{zhou2020cylinder3d} and PointNet++ \cite{Qi2017PointNetDH}, to capture both local and global geometric characteristics. The extracted features are subsequently passed through a multi-layer perceptron (MLP) for regression. To guarantee that the predicted covariance matrices are symmetric and positive definite, the network is designed to output a lower triangular matrix \( C_t \). The final covariance prediction is then reconstructed as:
\begin{equation}
    \widehat{\boldsymbol{Y}} = C_t C_t^\top.
\end{equation}
This formulation ensures that the predicted error covariance retains the necessary structural properties, namely, being positive semi-definite. . We also add a small diagonal term \(\varepsilon \mathbf{I}\) to make it strictly positive definite and numerically stable.

\subsubsection{Sampling Strategy and Data Augmentation}  
Due to the prevalence of samples with near-zero covariance in the dataset (Fig. \ref{covariance-on-point_cloud}), a weighted sampling strategy is employed during training. Specifically, the probability of selecting a sample for the input batch of our network is proportional to the magnitude of its largest covariance element, thereby emphasizing samples with higher error covariance. 
To further enhance model robustness, we apply a data augmentation strategy focused on rotational transformations. Each reference point cloud is randomly rotated about the vertical (\( z \)) axis to simulate various orientations. Following a rotation, the corresponding covariance matrices are adjusted using the adjoint representation of the transformation:
\begin{equation}
 \overline{\boldsymbol{Y}} = \mathbf{A}_{\boldsymbol{T}}\, \boldsymbol{Y}\, \mathbf{A}_{\boldsymbol{T}}^{\top},
\end{equation}
where \(\mathbf{A}_{\boldsymbol{T}}\) is the adjoint matrix associated with the applied rotation \(\boldsymbol{T}\). This augmentation increases the diversity of the training data while preserving the underlying geometric structure of the environment.

\subsubsection{Loss Function}\label{loss}

The training objective is formulated as a regression problem aimed at minimizing the difference between the predicted and ground truth covariance matrices. The loss function is defined as:
\begin{equation}
\begin{split}
\mathcal{L}(\widehat{\boldsymbol{Y}}, \overline{\boldsymbol{Y}}) =\; & \alpha \, D_{KL}\Bigl[\mathcal{N}\Bigl(0, \widehat{\boldsymbol{Y}}\Bigr) \parallel \mathcal{N}\Bigl(0, \overline{\boldsymbol{Y}}\Bigr)\Bigr] \\
& + \beta \, \sum_{i=1}^{n} L_{\text{Huber}}\Bigl(\widehat{\boldsymbol{Y}} - \overline{\boldsymbol{Y}}\Bigr),
\end{split}
\end{equation}
where \(\alpha = 0.01\) and \(\beta = 1.0\) are empirically determined hyperparameters, \(\widehat{\boldsymbol{Y}}\) is the predicted covariance matrix, and \(\overline{\boldsymbol{Y}}\) is the ground truth. The Huber loss is defined as:
\begin{equation}
L_{\text{Huber}}(\delta) =
\begin{cases}
\frac{1}{2} \delta^2, & |\delta| \leq \delta_0,\\[1ex]
\delta_0 \left(|\delta| - \frac{1}{2} \delta_0\right), & |\delta| > \delta_0.
\end{cases}
\end{equation}
Because the covariance matrix is symmetric, the Huber loss is applied only to its upper (or equivalently, lower) triangular elements.

\section{EXPERIMENT AND EVALUATION}
\subsection{Dataset}

Our proposed method is designed to be generalizable and applicable to any LiDAR-based dataset for localization. In this work, however, we conduct our experiments using the KITTI Odometry dataset \cite{geiger2012we}, which comprises 11 sequences with ground truth 6DoF poses. We extracted a point cloud every 50 scans (approximately every 5 seconds) from each sequence and followed Algorithm \ref{alg-dataset} to construct the dataset. For evaluation purposes, each sequence was partitioned into three consecutive segments with a split ratio of 70:20:10 for training, testing, and evaluation, respectively. For the creation of the reference map (e.g., \(x\) and \(y\) in Algorithm \ref{alg-dataset}), we explored two scenarios:
\begin{enumerate}
    \item \textbf{Scenario 1 (Pre-built maps):} Utilizes 10 preceding frames and 20 subsequent frames, emulating scenarios where a high-definition (HD) pre-built map is available for vehicle localization.
    \item \textbf{Scenario 2 (SLAM):} Utilizes only 10 preceding frames, relying exclusively on historical data, making it appropriate for simultaneous localization and mapping applications.
\end{enumerate}

The perturbation covariance matrix (\( \boldsymbol{\xi_o} \) in Algorithm \ref{alg-dataset}) is a critical component of our approach. The perturbation covariance matrix is parameterized by its diagonal elements, which represent the uncertainties of the initial transformations. These parameters are summarized in Table~\ref{tab:cov_params}.

\begin{table}[t]
\centering
\caption{The perturbation Covariance Parameters for generating datasets}
\label{tab:cov_params}
\begin{tabular}{ccccccc}
\hline
Parameter & \(\sigma_x\) [m] & \(\sigma_y\) [m]& \(\sigma_z\) [m] & \(\sigma_\phi\) [\textdegree] & \(\sigma_\theta\) [\textdegree] & \(\sigma_\psi\) [\textdegree] \\ \hline
Value     & 1.0             & 1.0            & 0.2          & 5.0               & 5.0               & 10.0             \\ \hline
\end{tabular}
\end{table}

\begin{figure}[t]
      \centering
      \includegraphics[width=\linewidth]{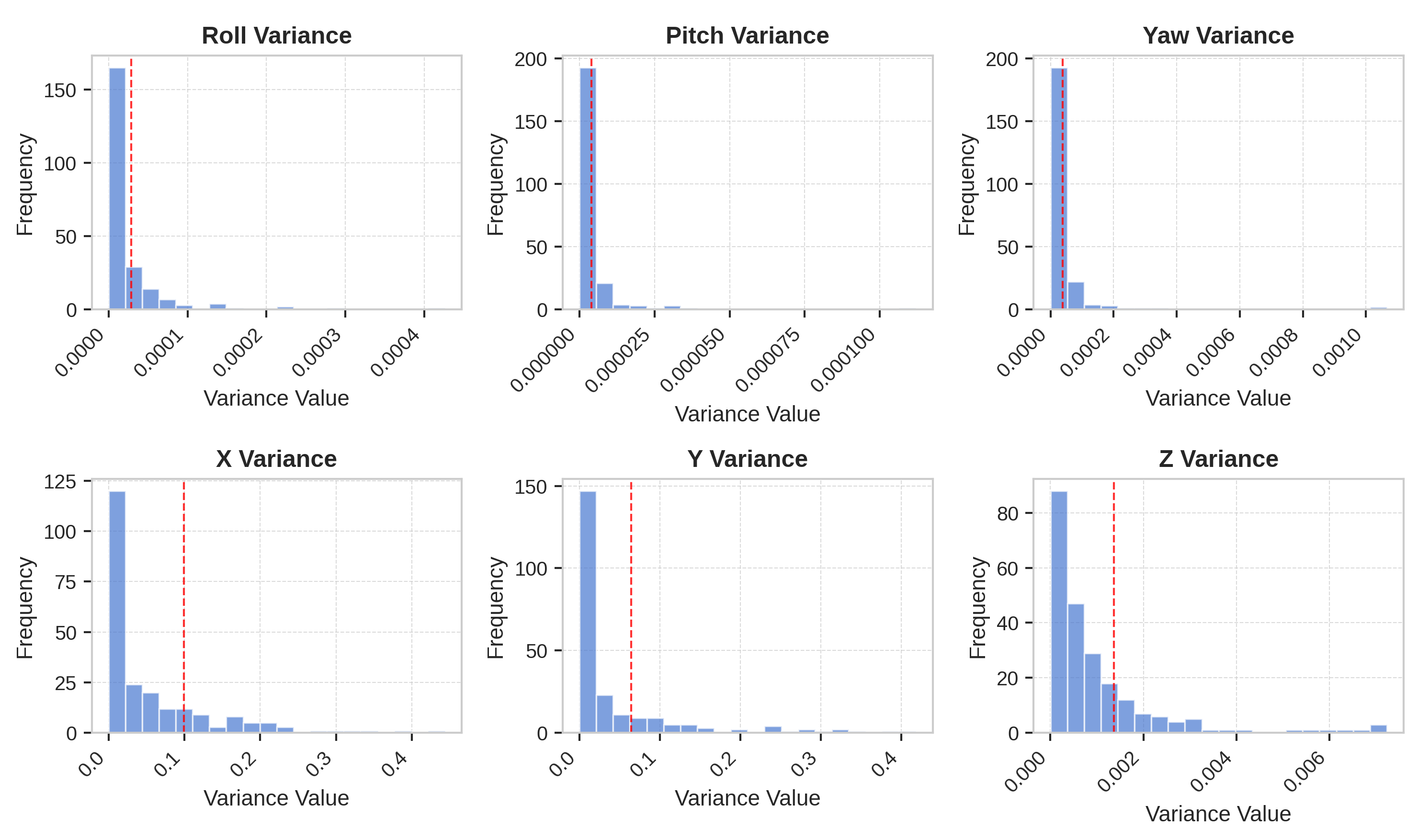}
      \caption{Histogram of localization error variance for sequence 00 of the kitti dataset, generated with Algorithm \ref{alg-dataset}. The red line indicates the 80\% percentile.}. 
      \label{covariance-on-point_cloud}
    \end{figure}

In Fig.~\ref{covariance-on-point_cloud}, the histograms depict the localization error variance across various dimensions for a sequence of the kitti dataset. As expected, most points show low error covariance, while the variance is most pronounced along the x-axis (indicating the vehicle's forward direction) and the y-axis (indicating the vehicle's leftward direction).

\begin{figure*}[!h]
    \centering
    \includegraphics[width=\textwidth]{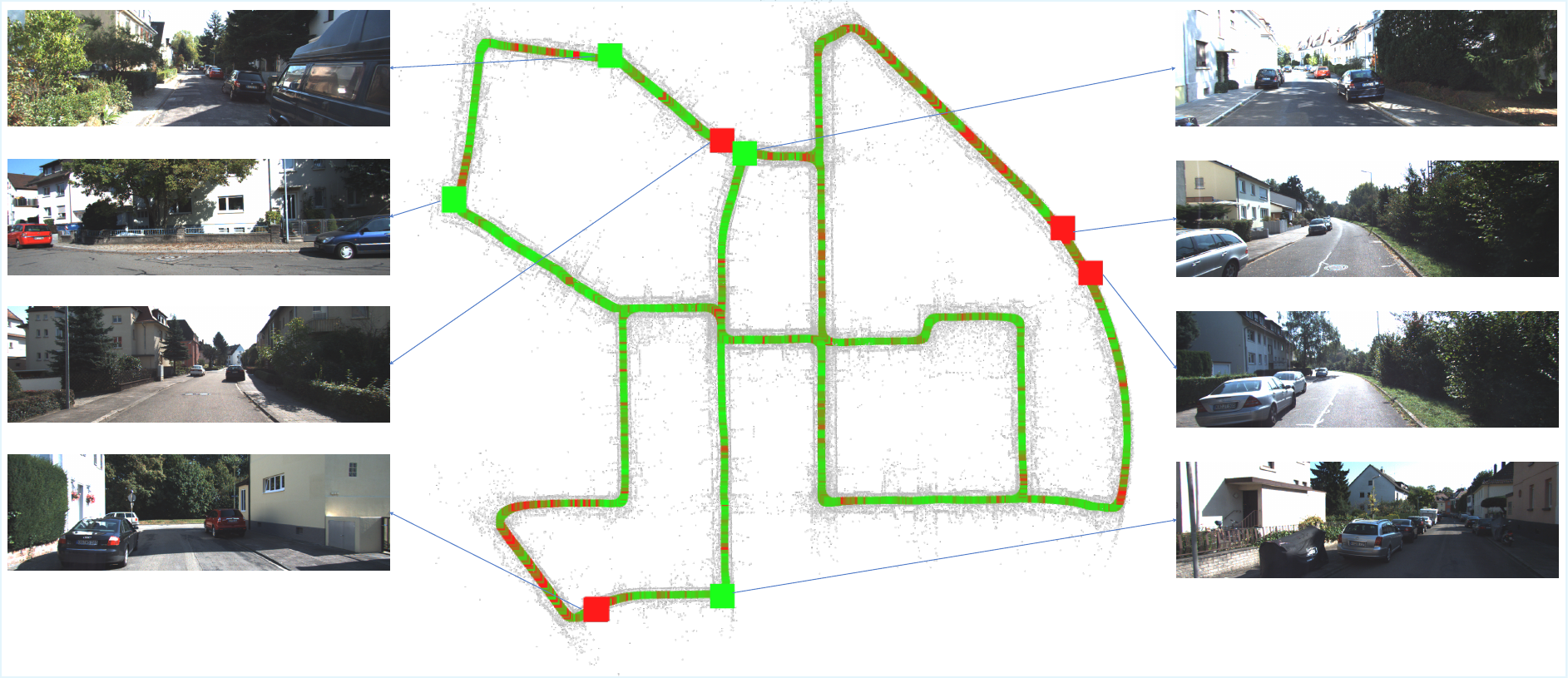}
    \caption{The map is reconstructed from consecutive LiDAR scans in KITTI Odometry Sequence~00. The trajectory is color-coded by $\operatorname{tr}(\Sigma_{xy})=\sigma_x^2+\sigma_y^2$ (green low, red high). As expected, we observe smaller error covariance at turns and crossroads, and larger covariance along straight segments.}
    \label{preds_seq00}
\end{figure*}

\begin{table}[t]
\scriptsize
\caption{Evaluation Metrics for Different Network Configurations and Dataset Strategies}
\begin{adjustbox}{max width=\textwidth}
\begin{tabular}{|c|c|c|c|c|}
\hline
\textbf{Network} & \textbf{KL} & \textbf{x (MAE)} & \textbf{y (MAE)} & \textbf{yaw (MAE)} \\ \hline
\multicolumn{5}{|c|}{\textbf{Localization}} \\ \hline
PointNet++ & \textbf{3.69}  & \textbf{0.059} & \textbf{0.041} & \(\boldsymbol{3.21 \times 10^{-5}}\) \\ \hline
Cylinder3D & 4.06 & 0.117 & 0.071 & \(5.50 \times 10^{-5}\) \\ \hline
Cylinder3D (pretrained) & 3.97 & 0.061 & 0.051 & \(5.36 \times 10^{-5}\) \\ \hline
\multicolumn{5}{|c|}{\textbf{Slam}} \\ \hline
PointNet++ & \textbf{4.28} & \textbf{0.063} & \textbf{0.035} & \(\boldsymbol{2.55 \times 10^{-5}}\) \\ \hline
Cylinder3D & 4.70 & 0.072 & 0.055 & \(5.52 \times 10^{-5}\) \\ \hline
Cylinder3D (pretrained) & 4.32 & 0.084 & 0.053 & \(5.441 \times 10^{-5}\) \\ \hline
\end{tabular}
\end{adjustbox}
\label{table:results}
\end{table}

\subsection{Results for Covariance Estimation}

Table \ref{table:results} presents the evaluation metrics for various network configurations and dataset strategies. The results indicate that PointNet++ outperforms the other models, particularly in the localization setting, where it achieves the most accurate covariance estimations with lower errors across all evaluated metrics. This suggests that its hierarchical feature extraction mechanism effectively captures relevant geometric information for uncertainty estimation. However, its performance slightly deteriorates in the SLAM setting, where the absence of subsequent frames impacts the estimation accuracy. 

In contrast, Cylinder3D exhibits higher errors across all settings, suggesting that it is less effective in capturing precise covariance estimations. One possible explanation is that Cylinder3D's voxel-based representation may lead to information loss in the context of fine-grained error estimation, which is crucial for accurate pose uncertainty prediction. Notably, the pretrained Cylinder3D model—which was trained on a point cloud segmentation task—demonstrates an improvement over its standard counterpart. This improvement highlights the potential benefits of pretraining on structured point cloud tasks, yet the model still performs sub-optimally compared to PointNet++. These findings suggest that further research is required to explore more effective pretraining strategies that can better transfer domain-specific knowledge to error covariance estimation.

In addition to these results, we visualize the predicted covariance along a single representative sequence of the KITTI dataset(Fig.~\ref{preds_seq00}). The map shows the vehicle’s trajectory, with each pose color-coded by the trace of the \(x\!-\!y\) position-covariance submatrix, \(\operatorname{tr}(\Sigma_{xy})=\sigma_x^2+\sigma_y^2\) (green = low, red = high). Empirically, red segments appear on straight right-of-way portions of the route, whereas turns tend to appear green—likely because turning increases viewpoint diversity and geometric constraints (parallax), improving observability and lowering \(\operatorname{tr}(\Sigma_{xy})\). This visualization allows us to pinpoint exactly where the model's performance degrades.

\begin{figure}
  \centering
  \includegraphics[width=\linewidth]{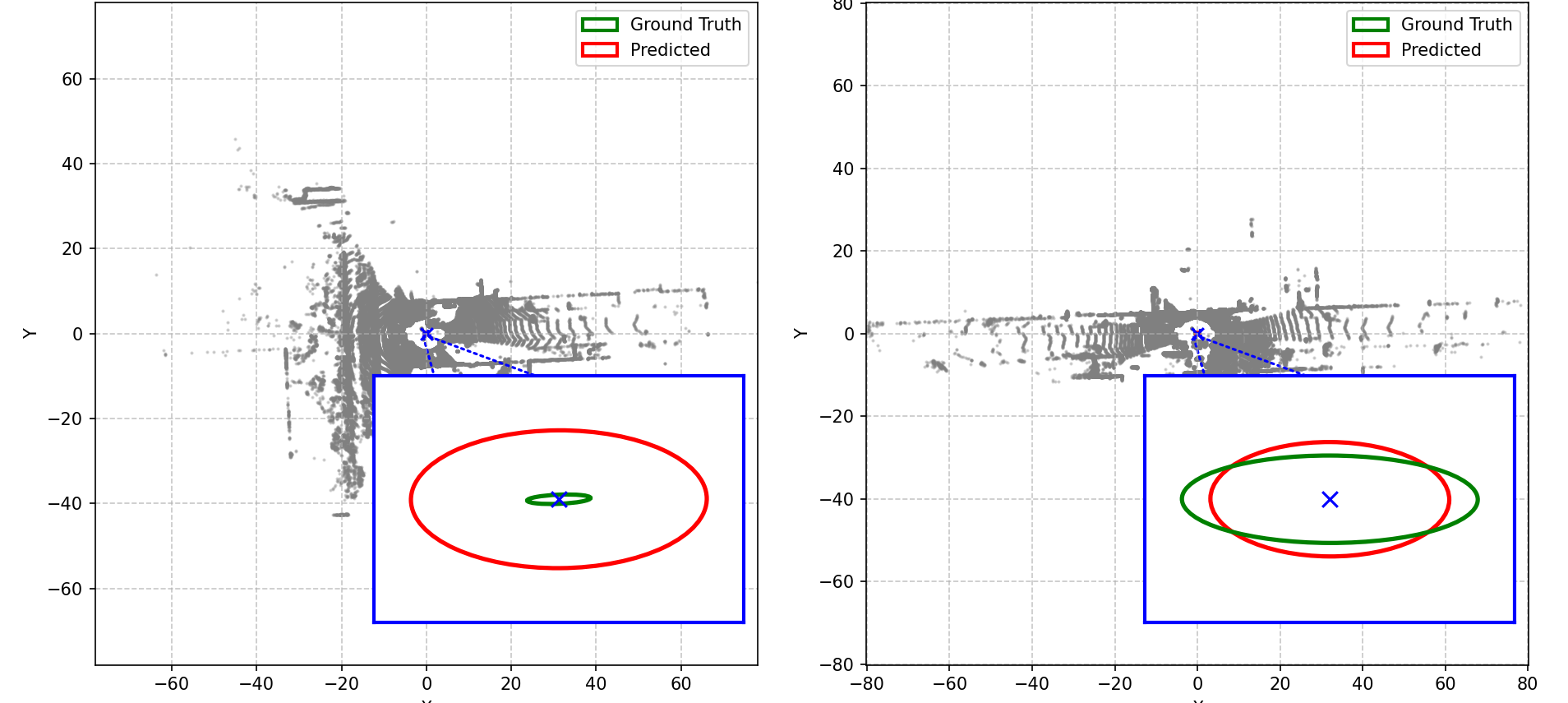}
  \caption{Two different predication of our best model in comparison to GT error covariance in x and y directions.}
  \label{vis_preds}
\end{figure}

To further analyze the performance of our models, we visually examined the predicted covariance values by plotting two representative outputs in Fig.~\ref{vis_preds}: one with a low KL divergence and one with a high KL divergence. Interestingly, our model tends to predict an uncorrelated covariance between the $x$ and $y$ directions, with generally higher values along the vehicle trajectory ($x$-direction). This observation is consistent with real-world scenarios, where uncertainty along the direction of motion is typically greater.

\subsection{Kalman–Filter Integration}

We smooth raw ICP poses with a constant–acceleration EKF (translation) and constant–angular–acceleration prior (rotation), without inertial sensing. The nominal state maintains rotation \(\mathbf{R}_k\in SO(3)\) and a 15-vector \([\mathbf{p}_k,\ \mathbf{v}_k,\ \mathbf{a}_k,\ \boldsymbol{\omega}_k,\ \dot{\boldsymbol{\omega}}_k]\). 

Between consecutive scans, we propagate the state using the scan time step and the constant-acceleration/constant-angular-acceleration priors, yielding a pose prior \(\mathbf{T}_k^{-}\). 
Each scan provides a 6-DoF ICP pose \(\widehat{\mathbf{T}}^{\mathrm{ICP}}_k\). We form a pose residual on \(SE(3)\) and apply a standard first-order EKF update around \(\mathbf{T}_k^{-}\).

Our network outputs a per-scan covariance \(\widehat{\boldsymbol{Y}}_k\in\mathbb{R}^{6\times6}\) \emph{before} matching. For temporal stabilization, we replace it with a 5-frame moving average
\(\widetilde{\boldsymbol{Y}}_k\) (last five predictions), and use this as the measurement noise.
As covariances are expressed in a local/sensor frame, we map them to the world frame using the \(SE(3)\) adjoint before using.

\begin{table}
\centering
\caption{Per-sequence mean improvement (\%) of per-frame covariance over a fixed baseline on KITTI Odometry.}
\label{tab:perseq_improvements}
\begin{adjustbox}{max width=\linewidth}
\begin{tabular}{@{} c c S[table-format=+2.2] S[table-format=+2.2] @{}}
\toprule
\textbf{Seq.\ \#} & \textbf{Split} & {\textbf{Avg APE Improvement} (\%)} & {\textbf{Avg RPE Improvement} (\%)} \\
\midrule
00 & val/test &  0.52 &  2.17 \\
01 & val/test & 10.60 &  1.92 \\
02 & val/test &  3.64 &  4.61 \\
04 & val/test &  2.71 &  1.45 \\
05 & val/test & 0.02 &  3.79 \\
06 & val/test &  0.64 & 1.74 \\
07 & val/test &  0.71 &  9.26 \\
08 & val/test & -0.19 & -0.02 \\
09 & val/test &  0.49 &  2.81 \\
10 & val/test &  1.13 &  0.13 \\
\bottomrule
\end{tabular}
\end{adjustbox}
\end{table}

\noindent\textbf{Evaluation protocol.}
We compare our \emph{per-frame} covariance weighting against a \emph{fixed} covariance baseline on three KITTI odometry sequences. For a fair comparison, the fixed baseline uses the \emph{per-sequence mean} of our predicted covariances as its constant measurement noise, and all Kalman filter hyperparameters are tuned once on this baseline and then reused \emph{unchanged} for our method. We operate in a pure LiDAR SLAM regime (no global sensors, no pose-graph optimization, no loop closure). Each sequence is partitioned into non-overlapping 200-frame windows; metrics are computed per window and then averaged. Because the first 70\% of each sequence is used to train the covariance predictor, we report \emph{val/test}, and present mean Absolute Pose Error (APE) \%  and Relative Pose Error (RPE) \% improvements over the fixed baseline.

As shown in Table~\ref{tab:perseq_improvements}, across the 10 KITTI Odometry sequences, our per-frame covariance generally improves or equals the fixed baseline on the held-out (val/test) splits for both APE and RPE. Training-split gains (not shown) are of similar magnitude, indicating the predictor is not overfitting to sequences it has seen.
 Overall, replacing a static noise model with our learned, scene-aware covariance yields cleaner trajectories without extra sensors or loop closure.

\section{CONCLUSION}
In this work, we introduced a deep learning-based framework for predicting the registration error covariance of the Iterative Closest Point (ICP) algorithm in LiDAR-based localization and SLAM systems. Our method directly estimates uncertainty from raw point clouds, improving sensor fusion through more accurate covariance modeling in Kalman filtering.
Through extensive experiments on the KITTI-Odometry dataset, we demonstrated that our approach enhances localization accuracy. Our visual analysis further highlighted the model’s tendency to predict uncorrelated covariances along the vehicle trajectory, suggesting the need for a more diverse dataset to refine estimation accuracy.
Moreover, integrating our predicted covariance into a Kalman filter demonstrated notable improvements in trajectory estimation. Our model outperformed a fixed covariance approach in both APE and RPE.
Future research will focus on optimizing network architectures, exploring alternative pretraining methods, and increasing dataset diversity to improve model robustness. Additionally, real-time deployment and multimodal sensor fusion, incorporating data from cameras and radar, will be investigated to further enhance localization accuracy in complex environments.

\bibliographystyle{IEEEtran}
\bibliography{sample-base}
\end{document}